\documentclass[sort&compress, numafflabel]{elsarticle}

\usepackage[]{natbib}
\usepackage{graphicx} 
\usepackage{subfiles}
\usepackage{multirow}
\usepackage{geometry}
\usepackage{url}
\usepackage{hyperref}
\usepackage{booktabs}
\usepackage{verbatim}
\usepackage{amsmath}

\newif\ifsubfile
\subfiletrue

\title{Health Text Simplification: An Annotated Corpus for Digestive Cancer Education and Novel Strategies for Reinforcement Learning}

\author{Md Mushfiqur Rahman\textsuperscript{*}, Mohammad Sabik Irbaz\textsuperscript{*}, Kai North, Michelle S. Williams \\ Marcos Zampieri, Kevin Lybarger}

\address{George Mason University}

\begin{document}

\begin{abstract}

\textit{Objective:} The reading level of health educational materials significantly influences the understandability and accessibility of the information, particularly for minoritized populations. Many patient educational resources surpass widely accepted standards for reading level and complexity. There is a critical need for high-performing text simplification models for health information to enhance dissemination and literacy. This need is particularly acute in cancer education, where effective prevention and screening education can substantially reduce morbidity and mortality.

\textit{Methods:} We introduce \textit{Simplified Digestive Cancer} (SimpleDC), a parallel corpus of cancer education materials tailored for health text simplification research, comprising educational content from the American Cancer Society, Centers for Disease Control and Prevention, and National Cancer Institute. The corpus includes 31 web pages with the corresponding manually simplified versions. It consists of 1,183 annotated sentence pairs (361 train, 294 development, and 528 test). Utilizing SimpleDC and the existing Med-EASi corpus, we explore Large Language Model (LLM)-based simplification methods, including fine-tuning, reinforcement learning (RL), reinforcement learning with human feedback (RLHF), domain adaptation, and prompt-based approaches. Our experimentation encompasses Llama 2, Llama 3, and GPT-4. We introduce a novel RLHF reward function featuring a lightweight model adept at distinguishing between original and simplified texts when enables training on unlabeled data.

\textit{Results:} Fine-tuned Llama models demonstrated high performance across various metrics. Our RLHF reward function outperformed existing RL text simplification reward functions. The results underscore that RL/RLHF can achieve performance comparable to fine-tuning and improve the performance of fine-tuned models. Additionally, these methods effectively adapt out-of-domain text simplification models to a target domain. The best-performing RL-enhanced Llama models outperformed GPT-4 in both automatic metrics and manual evaluation by subject matter experts.

\textit{Conclusion:} The newly developed SimpleDC corpus will serve as a valuable asset to the research community, particularly in patient education simplification. The RL/RLHF methodologies presented herein enable effective training of simplification models on unlabeled text and the utilization of out-of-domain simplification corpora.

\end{abstract}


\maketitle

\begin{center}
    \textsuperscript{*}Authors contributed equally to this paper.
\end{center}

\subfilefalse

\section{Introduction}

Accessibility and comprehensibility of patient medical educational materials are critical to improving health literacy and patient health \cite{kindig2004health}. Patient educational materials, including cancer-related materials, are essential to provide patients with the information necessary to make informed healthcare decisions. Health literacy is strongly influenced by the clarity and understandability of such materials and plays a key role in health outcomes \cite{cline2001consumer}. It affects various aspects of patient health, including reducing risk factors, encouraging timely screening, and facilitating proactive health management. To ensure comprehensibility for a broad audience, the American Medical Association (AMA) and the National Institutes of Health (NIH) recommend a reading level of sixth grade or lower for patient educational materials \cite{doi:10.1177/1090198105277329, doi:10.2214/AJR.13.11223}. Despite the critical role of these educational materials in patient health, there remains a substantial discrepancy between the recommended and actual readability observed in practice \cite{jindal2017assessing}. Health educational materials from prominent sources are often at high school or college reading levels \cite{jindal2017assessing}. This reading complexity poses a barrier, particularly for people with lower literacy or limited health knowledge, preventing them from fully understanding and taking action on essential health information. The accessibility and comprehensibility of educational materials are particularly concerning for disseminating cancer information, where clear and understandable guidance impacts patient education and decision-making \cite{rimer1984informed, mills1999importance}. Our focus on text-based cancer educational materials stems from their pivotal role in patient education and the pressing need to improve their accessibility to diverse populations.

Advancements in Natural Language Processing (NLP) and Large Language Models (LLMs) offer a promising avenue to reconcile the disparity in reading levels in existing medical education materials through automatic simplification, to improve public health literacy \cite{van2019evaluating}. LLMs have demonstrated proficiency in reducing linguistic complexity and preserving medical meaning \cite{devaraj2021paragraph, feng2023sentence}. Finding the optimum balance of simplification and information retention is challenging, especially when the stakes of miscommunication are high \cite{mccabe2018miscommunication}. Developing and evaluating data-driven text simplification models requires parallel corpora of original and simplified text. Creating these parallel corpora is extremely labor-intensive and costly. In the health domain, these challenges are exacerbated by the need for domain-specific medical expertise. To address these challenges, robust methods are needed to create high-performing simplification models with minimal training data.

This paper makes significant contributions to the simplification of health information and text simplification more broadly. We focus on patient education materials related to digestive cancer, given the prevalence and mortality of these cancers \cite{rawla2019epidemiology, teglia2023association}. We introduce \textit{Simplified Digestive Cancer (SimpleDC)}, a novel text simplification corpus comprising patient educational materials from three prominent sources: National Cancer Institute (NCI), American Cancer Society (ACS), and Centers for Disease Control and Prevention (CDC). Although prior research has explored the simplification of health information, we are unaware of prior research focused on patient educational materials. SimpleDC is unique in its concentration on digestive cancers and includes parallel sentences (original and simplified versions), where simplifications were generated by a team of nurse oncologists and a nurse practitioner with medical and patient education backgrounds. We explore supervised fine-tuning (SFT) and reinforcement learning (RL) approaches using Llama 2 \cite{touvron2023llama2} and Llama 3 \cite{llama3modelcard} and introduce a novel approach for reinforcement learning from human feedback (RLHF). We also explore prompt-based methods using GPT-4 \cite{Achiam2023GPT4TR, brown2020language} and introduce a novel self-correction strategy that utilizes emergent reasoning capabilities. Our evaluation includes standard automatic metrics (e.g. SARI, BLEU, BERTScore, and FKGL) and a human comparison of LLM and human-generated simplifications. Automatic and manual evaluations demonstrate that the proposed RLHF reward function outperforms existing text simplification reward functions, complements SFT, and facilitates domain adaptation. Our findings advance the simplification of health text and enable the automatic creation of more accessible health information, ultimately contributing to improved health literacy. We provide the new SimpleDC corpus, trained models, and code to the research community.\footnote{Link to dataset: \href{https://github.com/mushfiqur11/simpledc-dataset.git}{https://github.com/mushfiqur11/simpledc-dataset.git}} \footnote{Link to code: \href{https://github.com/mushfiqur11/healthLiteracy.git}{https://github.com/mushfiqur11/healthLiteracy.git}}\\

\noindent\textbf{Statement of Significance}
\begin{table}[h]
    \centering
    
    \begin{tabular}{lp{0.7\linewidth}}
    \toprule
        Problem & Existing patient education materials are frequently presented at a high reading level, reducing accessibility for individuals with lower literacy. \\
        \midrule
        What is already known & Existing health text simplification corpora are limited, and there are knowledge gaps for low-resource text simplification related to RL, RLHF, and LLMs. \\
        \midrule
        What this paper adds & We present a novel text simplification corpus, SimpleDC, and introduce an innovative RLHF reward function for training LLM simplification models. The findings demonstrate the RLHF approach surpasses existing methods in performance, complements supervised fine-tuning, and supports domain adaptation. \\
        \bottomrule
    \end{tabular}
\end{table}

\ifsubfile
\bibliography{mybib}
\fi

\section{Related Work}

Text simplification involves the transformation of texts to improve understandability and accessibility to broad audiences. This section presents the scope of text simplification, health text simplification corpora, text simplification methods, and related RL research.

\subsection{Scope of Text Simplification}

Text simplification has been explored in many domains (news articles, literature, scientific articles, etc.) where complex ideas must be communicated clearly and concisely \cite{cao-etal-2020-expertise, nisioi-etal-2017-exploring, maddela-etal-2021-controllable, LCPsurvey, shardlow-etal-2020-complex}. This research spans multiple languages, addressing the unique challenges and strategies needed for multilingual simplification to ensure clarity for linguistically diverse audiences \cite{ryan2023revisiting, koptient2020fine, stajner-etal-2015-automatic}. Text simplification can be subdivided into three tasks: lexical simplification, sentence-level simplification, and paragraph/document-level simplification \cite{grabar-saggion-2022-evaluation, shardlow2014survey, north2023deep}. Lexical simplification commonly involves substituting words or phrases, including medical terminology \cite{ramadier2018radiological, ZilioLeonardo2020ALST, qenam2017text, kandula2010semantic, leroy2013user}. However, word- or phrase-level editing limits simplification scope and cannot accommodate more comprehensive rephrasing or restructuring. Sentence- and document-level simplification enables more comprehensive text restructuring to enhance overall clarity and accessibility and typically requires parallel text corpora (complex and simplified text) for model training and evaluation \cite{lu2023napss}. 

\subsection{Health Text Simplification}

Health text simplification is challenging due to medical jargon, complex sentence structures, and lengthy explanations. It has been explored through several tasks, such as generating patient-friendly summaries of clinical notes \cite{liang-etal-2019-novel-system, kanwal2022attention, joshua2011summarization}, making biomedical literature accessible to laypeople, and simplifying educational materials for patients \cite{Guo_Qiu_Wang_Cohen_2021, abrahamsson-etal-2014-medical, van2019evaluating, van-etal-2020-automets, cardon-grabar-2019-parallel, sakakini-etal-2020-context}. Text simplification can focus on improving readability or making stylistic modifications; however, simplification of health information requires rigorous attention to meaning preservation and medical accuracy, given the critical implications of miscommunication \cite{van2019evaluating, kandula2010semantic}. This challenge is amplified by the diverse needs of patients, which encompasses varying levels of literacy, language background, and other factors that require a nuanced approach to achieve clarity and accuracy.

Due to the labor-intensive nature of annotation, there are limited health-focused text simplification corpora \cite{medeasi-basu, phatak2022medical, devaraj2021paragraph, van-etal-2020-automets}. The Medical dataset for Elaborative and Abstractive Simplification (Med-EASi) dataset comprises medical text from SIMPWIKI that was simplified by a team of medical experts and layperson crowd workers at the sentence-level \cite{medeasi-basu}. The Autocomplete for Medical Text Simplification (AutoMeTS) dataset consists of automatically aligned sentences from English Wikipedia and Simple English Wikipedia for medical topics \cite{van-etal-2020-automets}. Devaraj et al. provided a manually annotated parallel corpus of biomedical literature with simplified text created by domain experts \cite{devaraj2021paragraph}. We introduce a new health text simplification corpus, SimpleDC, which is based on educational patient materials from prominent information sources (ACS, CDC, and NCI) and explicitly focuses on the cancer domain.

\subsection{Text Simplification Approaches}

Early text simplification work used heuristics and traditional machine learning techniques, including pattern-based paraphrasing \cite{filippova-strube-2008-dependency, filippova-strube-2008-sentence}. Contemporary approaches conceptualize text simplification similar to machine translation, employing neural language models to transform sentences into simpler counterparts \cite{van2019evaluating, shardlow-nawaz-2019-neural}. Document-level simplification adopts a similar machine translation approach \cite{sun-etal-2021-document,sun-etal-2022-rethinking}. Essentially all state-of-the-art text simplification models \cite{sheang-saggion-2021-controllable,farajidizaji2023possible} are based on LLMs, including: i) encoder-decoder architectures, like Text-to-Text Transfer Transformer (T5) \cite{raffel2020t5} and Fine-tuned LAnguage Net on T5 (FlanT5) \cite{chung2022flant5} and ii) decoder-only models such as Large Language Model Analysis (Llama) variants \cite{touvron2023llama, touvron2023llama2, llama3modelcard} and Generative Pre-trained Transformer (GPT) \cite{brown2020language} variants, like GPT-4 \cite{Achiam2023GPT4TR}. 

LLM simplification models are predominantly created through SFT; however, there has been some limited exploration of prompt-based methods and reinforcement learning. Instruction tuning allows LLMs to follow natural language instructions, facilitating prompt-based approaches such as in-context learning, chain-of-thought (CoT) learning \cite{wei2022chain, wang-etal-2023-self-prompted}, and self-correction \cite{wu2023self,pan2023automatically}. These methods rely on the inherent language understanding and reasoning capabilities of LLMs \cite{wei2022chain, liu2023summary, tian2024opportunities}. CoT learning enables models to articulate step-by-step reasoning, similar to human problem solving, improving response accuracy and interpretability \cite{wei2022chain, wang-etal-2023-self-prompted}. Self-correction allows models to iteratively refine their outputs, improving the precision and reliability of their responses \cite{pan2023automatically}. We present novel approaches for adapting these techniques to our text simplification task.

\subsection{Reinforcement Learning for Simplification}

RL has emerged as a promising approach for text simplification \cite{phatak2022medical, nakamachi-etal-2020-text, yanamoto-etal-2022-controllable}. Many studies employ a mix of simplicity metrics and linguistic features to construct their reward functions. Zhang et al. \cite{zhang-lapata-2017-sentence} introduced an RL approach, DRESS, which includes a reward function focused on simplicity, fluency, and meaning preservation. Nakamachi et al. explored the use of rewards related to grammaticality, meaning preservation, and simplicity \cite{nakamachi-etal-2020-text}. Similarly, Yanamoto et al. introduced an RL approach based on the difference between estimated and target difficulty levels, using a novel reward calculation method \cite{yanamoto-etal-2022-controllable}. Alkaldi et al presented an RL paradigm that refines reading levels in a controlled and iterative process \cite{alkaldi2023text}. Luo et al demonstrated the use of controllability in the biomedical sector \cite{luo2022readability}. The TESLEA RL framework integrates heuristic and data-driven reward models to develop LLM-based text simplification models \cite{phatak2022medical} using three rewards: 1) readability (FKGL), 2) relevance (cosine similarity between sentence embeddings), and 3) lexical simplicity (Zipf frequency). 
In our baseline experimentation, we implemented a TESLEA-inspired approach that includes the readability and relevance rewards but omits the Zipf frequency-based lexical simplicity reward to avoid penalizing medically significant but infrequent words. We introduce a new RL approach that combines the heuristic readability reward (FKGL) with a data-driven RLHF reward for assessing simplification.

\ifsubfile
\bibliography{mybib, Kai_LexicalSimp}
\fi

\section{Materials and Methods}

\subsection{Data}

\subsubsection{SimpleDC}

\label{sec:simpledc}

We introduce \textit{SimpleDC}, which is a meticulously curated parallel corpus, specifically designed for text simplification research within the medical domain. It comprises patient educational information from three high-reputation institutions: ACS, CDC, and NCI. These organizations were chosen because of their expertise and comprehensive coverage of health topics, including cancer. There are many types of cancer, each with different characteristics. The available cancer educational materials on the ACS, CDC, and NCI websites exceed our annotation budget, so we focused on the subset of cancers associated with the digestive system to create a corpus that spans many cancer types that share similar terminology and topics (e.g. anatomy or symptoms). The focus on digestive cancers is also motivated by the prevalence and mortality of these cancers and the role that prevention and screening play in the outcomes \cite{li2022digestive}. SimpleDC responds to the need to make medical information more accessible and comprehensible to a broad audience, especially minoritized populations.

\begin{table}[ht]
\small
\begin{tabular}{p{0.35in} p{2.0in} p{2.0in} p{0.4in} p{0.4in}}
\toprule
\textbf{Source}  & \textbf{Original Text} & \textbf{Simplified Text} & \textbf{Orig. FKGL} & \textbf{Simp. FKGL} \\ \midrule
ACS     & The anus is the opening at the lower end of the intestines.   & The anus is the opening where bowel movements come out of your body. & 4.4 & 3.6 \\ \midrule
CDC     & Screening can find precancerous polyps—abnormal growths in the colon or rectum—that can be removed before they turn into cancer. & Screening can help find growths in your colon or rectum called precancerous polyps. They can be removed before they turn into cancer. & 10.7 & 5.2 \\ \midrule
NCI     & Primary liver cancer is a disease in which malignant (cancer) cells form in the tissues of the liver. & A cancer that starts in your liver is called primary liver cancer. & 10.3 & 6.8 \\ \bottomrule
\end{tabular}
\caption{Annotation examples and Flesch-Kincaid Grade-Level (FKGL) scores}
\label{annotation_example}
\end{table}

We collected text from the ACS, CDC, and NCI web pages containing patient educational content for eight digestive cancers: anal, bile duct, colorectal, gastrointestinal, liver, pancreatic, small intestine, and stomach. For these cancer types, we collected the text from web pages presenting introductory cancer information (e.g. About, Prevention, Screening). The collected text was cleaned and processed, including the removal of extraneous white space, adjusting punctuation, removing rich-text formatting, and parsing into sentences. We developed annotation guidelines for the simplification task and trained a team of subject matter experts with expertise in both cancer and patient education, including two nurse oncologists and a nurse practitioner. For each web page, we created a two-column document, where the left and right sides contained the original text with each sentence on a separate line. The annotators modified the sentences in the right column based on the annotation guidelines to create the simplified versions. This approach allowed the annotators to focus on sentence-level simplification while considering the entire web page for context. Table \ref{annotation_example} presents annotation examples for each source, together with the Flesch-Kincaid Grade Level (FKGL) scores for each example \cite{kincaid1975derivation}. FKGL is a rule-based readability formula that assesses text complexity based on the average sentence length and syllable count per word, yielding scores indicative of the US school grade levels required for comprehension.

\begin{table}[ht!]
\small
\centering
\begin{tabular}{lrrrrrrc}
\toprule
 & \multicolumn{1}{l}{} & \multicolumn{1}{l}{} & \multicolumn{2}{c}{Original FKGL} & \multicolumn{2}{c}{Simplified FKGL} &  \\ 
 & \multicolumn{1}{c}{Num. samples} & \multicolumn{1}{c}{Num. of webpages} & \multicolumn{1}{c}{Mean} & \multicolumn{1}{c}{Median} & \multicolumn{1}{c}{Mean} & \multicolumn{1}{c}{Median} & Annotation type \\
\midrule
Train & 361 & 14 & 9.6 & 9.1 & 7.8 & 8.0 & Single \\
\midrule
Val & 294 & 7 & 7.7 & 7.2 & 7.1 & 7.0 & Double \\
\midrule
Test & 528 & 10 & 8.1 & 7.6 & 7.4 & 7.5 & Double \\
\midrule
Overall & 1183 & 31 & 8.5 & 8 & 7.4 & 7.6 & - \\
\bottomrule
\end{tabular}

\caption{Statistics of the proposed Simple DC dataset}
\label{simpledc_stats}

\end{table}

\begin{figure}[ht!]
    \centering
    \includegraphics[width=0.9\linewidth]{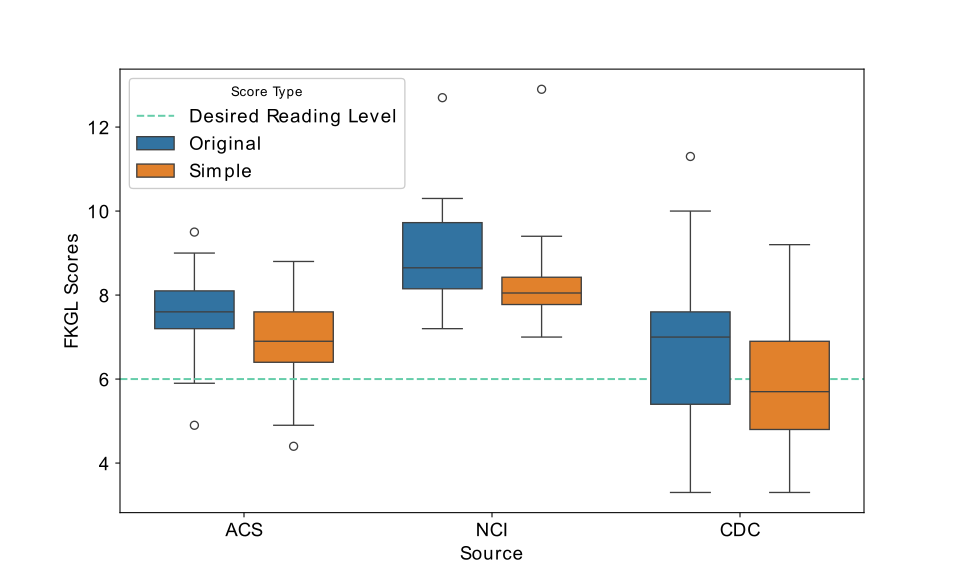}
    \caption{Boxplot representing the FKGL scores of Original and Simple Text of the SimpleDC dataset (grouped by source)}
    \label{fkgl_boxplots}
\end{figure}

Since text simplification is a generative task without categorical labels, traditional inter-annotator agreement (IAA) metrics, like Cohen’s Kappa, are not well suited. Instead, we evaluated annotation quality by having two annotators create the simplified text and then having a third annotator assess their preference for simplifications. In the initial training round, two annotators independently generated simplifications for each web page. The generated texts from the two annotators were marked \textit{A} and \textit{B} randomly. To assess annotation quality, a third annotator reviewed the simplifications relative to the original text and indicated their preferred simplification as one of four choices -- \textit{A} is better; \textit{B} is better; \textit{Both} are good; or \textit{None} is good. The distribution of the third annotator's preferences was as follows: the first annotator was preferred for 34\% of samples, the second annotator was preferred for 32\% of samples, both annotators were considered effective for 26\% of samples, and neither simplification was deemed adequate for 8\% of samples. SimpleDC comprises 31 annotated web pages (14 train, 7 development, and 10 test), including 1,183 annotated sentence pairs (361 train, 294 development, and 528 test). The development and test sets were doubly annotated by two annotators and adjudicated by the third annotator to create a robust evaluation set. The training set was singly annotated. Figure \ref{simpledc_stats} gives a detailed overview of the SimpleDC dataset. Figure \ref{fkgl_boxplots} presents the distribution of the FKGL reading level for the original and simplified text for different data sources. The median reading level varies 7.0-9.2 across sources, and the reduction in FKGL from simplification ranged 0.6-1.0.

We collected an additional 1,395 web pages from ACS, CDC, and NCI related to digestive cancer that were not annotated/simplified. We randomly sampled 1,000 of these unannotated web pages for use in the RL procedures described in Section \label{RL_methods}.

\subsubsection{Med-EASi}
Our experimentation utilized the Med-EASi \cite{medeasi-basu} corpus, which includes simplifications for medical texts annotated through a combination of expert, layperson, and AI-generated contributions. The more complex medical texts are annotated by experts, while simpler texts are handled by layperson crowd-workers, often assisted by AI-generated suggestions. It includes annotations for specific edit types: elaboration, replacement, deletion, and insertion. There are 1,979 original-simplified sentence pairs, covering a wide range of medical topics. The average FKGL of the original and simplified samples is approximately 13\textsuperscript{th} and 10\textsuperscript{th} grades, respectively. The introduction of Med-EASi includes simplification results based on fine-tuning T5-large \cite{raffel2020t5}. Experimentation included a conventional machine translation approach using T5, as well as a controlled simplification approach that utilized the edit types.

\subsection{Simplification Models}

Our exploration of SimpleDC and Med-EASi focused on state-of-the-art decoder-only LLMs, including Llama 2 \cite{touvron2023llama2}, Llama 3 \cite{llama3modelcard} and GPT-4 \cite{Achiam2023GPT4TR}. Using Llama models, we explored zero-shot, SFT, and RL/RLHF approaches. Using GPT-4, we explored prompt-based strategies, including a novel self-correction strategy.

\subsubsection{Supervised fine-tuning (SFT)}

As a baseline, we trained Llama models on original-simplified sentence pairs using conventional SFT with cross-entropy loss. SFT experimentation included SimpleDC and Med-EASi.

\begin{figure}[ht!]
    \centering
    \includegraphics[width=\linewidth]{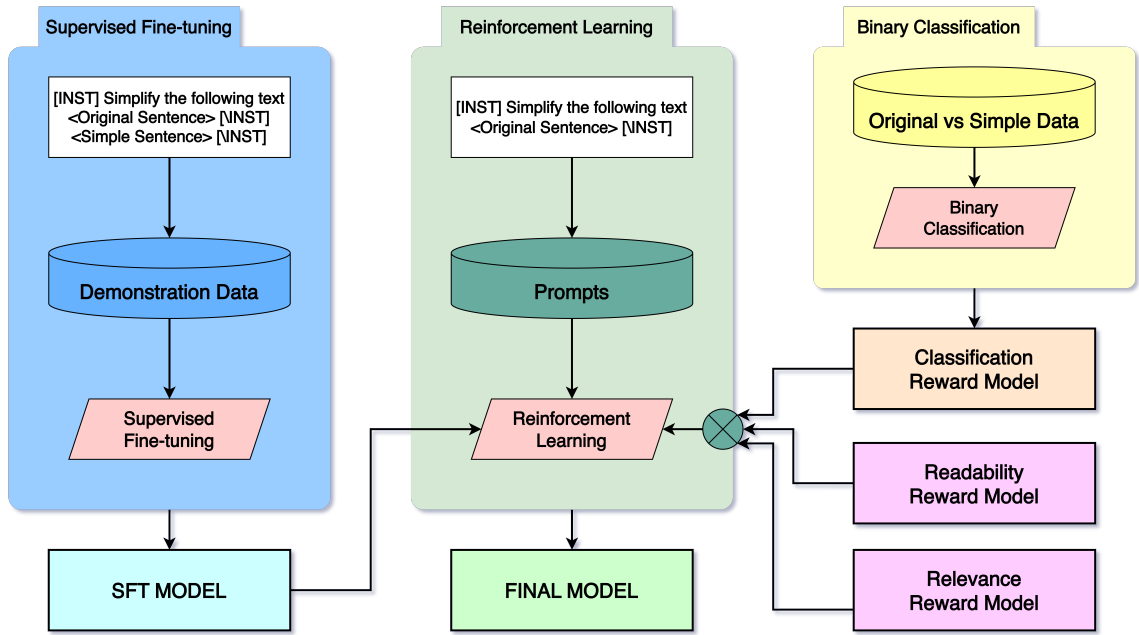}
    \caption{Reinforcement Learning from Human Feedback for Medical Text Simplification}
    \label{sft_rlhf}
\end{figure}

\subsubsection{Reinforcement Learning}
\label{RL_methods}

We explored RL strategies to incorporate unlabeled in-domain text into training. Using Proximal Policy Optimization (PPO) \cite{schulman2017proximal}, we investigated RL reward functions that include metrics related to reading level assessment (i.e., FKGL), topical relevance (semantic similarity), and a data-driven approach to identify simplified text (referred to as Original vs. Simplified below). This exploration includes RLHF \cite{NIPS2017_d5e2c0ad}. Figure \ref{sft_rlhf} presents an overview of the RL pipeline, each reward metric is described below. RL frameworks commonly use reward values in the range $[0, 1]$, where higher values indicate more desirable outcomes and lower values indicate less desirable outcomes. For each reward metric presented, individual and composite rewards are scaled to this range.


\textbf{Readability Reward:} FKGL is a heuristic function to assess reading level. Although an imperfect metric, FKGL provides an automatic way to assess the reading grade level of the text. The TESLEA approach \cite{phatak2022medical} includes a normalization strategy to convert FKGL scores to the range $[-1, 1]$, based on the target reading level. Our reading level reward, $R_{FKGL}$, which was inspired by the TESLEA approach, was calculated by normalizing the FKGL score by the reading level target and applying a sigmoid function:
$$ R_{FKGL} = sigmoid\left(\frac{6.5 - FKGL(generated)}{6.5}\right)$$
The normalized reading score (fraction within the sigmoid) yields positive values for simpler text (more desirable) and negative values for more complex text (less desirable). The sigmoid function maps the normalized reading scores to the range $[0, 1]$, where higher values indicate lower reading levels.

\textbf{Relevance Reward:} The relevance reward, $R_{Rel}$, is intended to quantify the semantic relationship between the simplified and original texts, where a higher reward indicates greater semantic similarity. We follow the TESLEA approach \cite{phatak2022medical} for relevance scoring: 1) the original and simplified texts are mapped to separate vector representations using an encoder and 2) semantic similarity is assessed based on cosine similarity between these vectors. BioSentVec \cite{chen2019biosentvec} was selected as the encoder, as it has been extensively trained on biomedical literature. Cosine similarity falls in the range $[0, 1]$, where higher values indicate higher semantic correspondence. The relevance reward incentivizes the preservation of semantic relatedness between the original and simplified texts. We used the pre-trained BioSentVec without any training on SimpleDC.

\textbf{Original vs. Simplified (OvS) Reward:} We introduce a new RL reward, $R_{OvS}$, based on the likelihood that a sample has been simplified. Using SimpleDC training data, we trained a binary classifier to label samples as original or simplified using BioMed-RoBERTa base \cite{gururangan-etal-2020-dont}. Similar to InstructGPT \cite{NEURIPS2022_b1efde53}, we utilize a classifier trained on human-annotated data; however, in our formulation, the classifier distinguishes between original and simplified texts. This classifier provides a probabilistic reward signal reflecting human judgment, capturing annotators' preferences and is considered RLHF. The \textit{OvS} classifier achieved 75.1\% accuracy on the SimpleDC development set. This performance demonstrates the ability to identify key features of the original and simplified texts. The reward value, $R_{OvS}$, is the predicted probability that the text is simplified ($P(simplified)$) providing a continuous reward signal ranging from $[0, 1]$, where higher values suggest better simplification. Our intuition behind this reward function is that it provides a mechanism for learning features of the original and simplified text and propagating this learning to unlabeled samples during RL.

\textbf{Reward Aggregation:} We calculated aggregated reward values from the individual metrics using the harmonic mean, $H$, to balance the contribution from each metric, as 
$$H(x_1, x_2) =\frac{1}{\frac{1}{x_1} + \frac{1}{x_2}}$$ 
\noindent
We developed two aggregated reward functions:
\begin{itemize}\setlength\itemsep{0em}
    \item $\mathbf{R_{FKGL+Rel}}$: We combined the reading level reward, $R_{FKGL}$, and the relevance reward, $R_{Rel}$, to create an aggregated reward, $R_{FKGL+Rel} = H(R_{FKGL}, R_{Rel})$. This reward is intended to capture both text complexity and semantic similarity to yield text that is at the desired reading level and preserves the original meaning. It does not incorporate any learning from SimpleDC and is therefore considered RL.
    \item $\mathbf{R_{FKGL+OvS}}$: We also combined the reading level reward, $R_{FKGL}$, and the Original vs. Simplified reward, $R_{OvS}$, to create an aggregated reward, $R_{FKGL+OvS}=H(R_{FKGL}, R_{OvS})$. This aggregated reward is intended to assess text complexity ($R_{FKGL}$) and identify linguistic features that differentiate the original and simplified texts ($R_{OvS}$) to produce text that is both at the desired reading level and emulates the simplification of SimpleDC. As this reward function incorporates a training signal derived from the SimpleDC annotations, it is considered RLHF.
\end{itemize}
In initial experimentation, we explored weighted averages, drawing inspiration from the TESLEA paper \cite{phatak2022medical}. However, due to inconsistent outcomes with hyperparameters and lower performance, we chose to focus on experiments employing the harmonic mean.

\subsubsection{Fine-tuning With Reward Models}

We explored different pipelines for incorporating RL/RLHF, including only using RL/RLHF, combining SFT and RL/RLHF, and using RL/RLHF for domain adaptation:
\begin{itemize}\setlength\itemsep{0em}
    \item \textit{RL/RLHF only}: A pre-trained LLM was trained using RL/RLHF with unlabeled in-domain data (digestive cancer text) to learn the target domain and simplification task without any SFT of the LLM on annotated in-domain data (SimpleDC). The goal of this exploration was to understand the achievable performance without any SFT.
    \item \textit{In-domain SFT + RL/RLHF}: A pre-trained LLM was first trained through SFT on labeled in-domain data (SimpleDC train set), followed by continued training through RL/RLHF on unlabeled in-domain data (digestive cancer text). This experimentation was intended to assess whether RL/RLHF can continue to improve the model even after in-domain SFT.
    \item \textit{Out-domain SFT + RL/RLHF}: A pre-trained LLM was first trained through SFT on labeled out-of-domain data (Med-EASi training set), followed by continued training through RL/RLHF unlabeled in-domain data (digestive cancer text).
\end{itemize}
In this experimentation, the unlabeled in-domain text consisted of 1,000 web pages from ACS, CDC, and NCI related to digestive cancer that were \textit{not annotated} and did not overlap with SimpleDC.

In all RL experiments, training used PPO, where a Kullback-Leibler (KL) divergence constraint  \cite{csiszar1975divergence} was implemented to limit excessive deviation of the RL-trained LLM from the original LLM. This divergence constraint provides a regulatory mechanism to ensure the LLM maintains linguistic coherence and avoids overfitting to potentially incorrect or excessive scores from the reward model. This constraint was essential due to the inherent limitations of the reward model, which may not accurately capture the quality of the generated text.

\subsubsection{Prompt-based models}

Given our focus on developing simplification models with limited training data and the success of GPT-4 in many tasks, we explored prompt-based approaches for text simplification with GPT-4 \cite{Achiam2023GPT4TR}, including in-context learning, CoT, and self-correction.

\begin{figure}[ht!]
    \centering
    \includegraphics[width=0.60\linewidth]{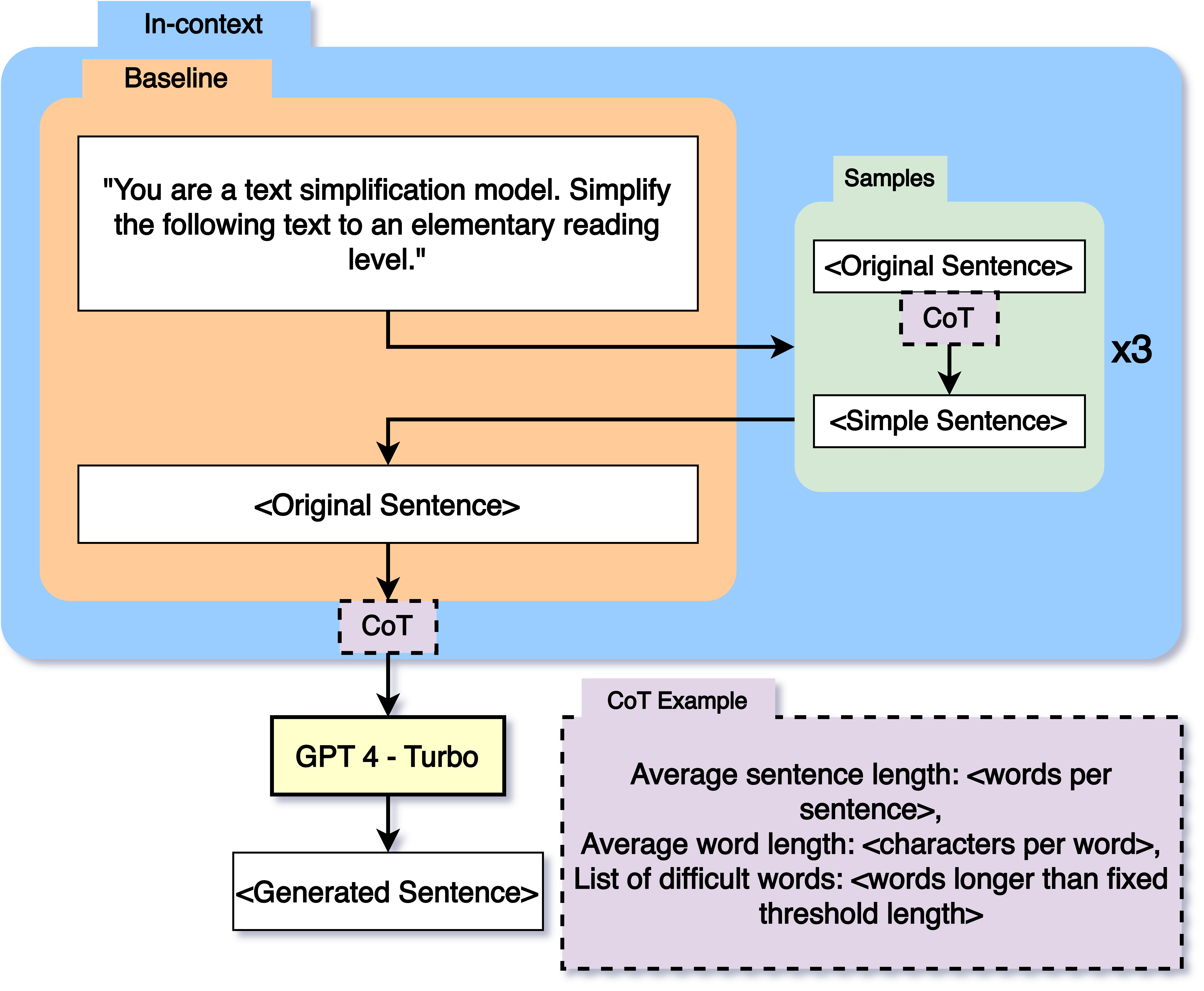}
    \caption{Chain-of-Thought (CoT) prompting along with Baseline and In-context}
    \label{cot_fig}
\end{figure}

\textbf{Zero-shot:} In the zero-shot approach, GPT-4 was instructed to simplify the text to an elementary level of reading, to provide a baseline for evaluating more complex prompting strategies. The specific simplification prompt is included in the \textit{Baseline} module in Figure \ref{cot_fig}.

\textbf{In-context Learning:} The in-context variant used the same initial prompt as the zero-shot approach but was augmented with three representative examples from the SimpleDC training set. These examples were manually selected to illustrate the desired simplification style and complexity.

\textbf{FKGL-Enhanced Prompt:} Building on the initial prompt, we added a detailed description of the FKGL heuristic, indicating that lower reading levels are associated with shorter words and sentences, and provided the FKGL formula for grade-level calculations.

\textbf{Chain-of-Thought (CoT):} Our CoT approach builds on the \textit{FKGL-Enhanced Prompt} approach by providing the model with additional sample-specific context, including the average word length, average sentence length, and a list of longer words ($\ge 3$ syllables) for each sample. This model aims to provide explicit guidance for the simplification process and anchor the simplification in commonly used readability standards. Figure \ref{cot_fig} illustrates the CoT approach.

\begin{figure}
    \centering
    \includegraphics[width=\linewidth]{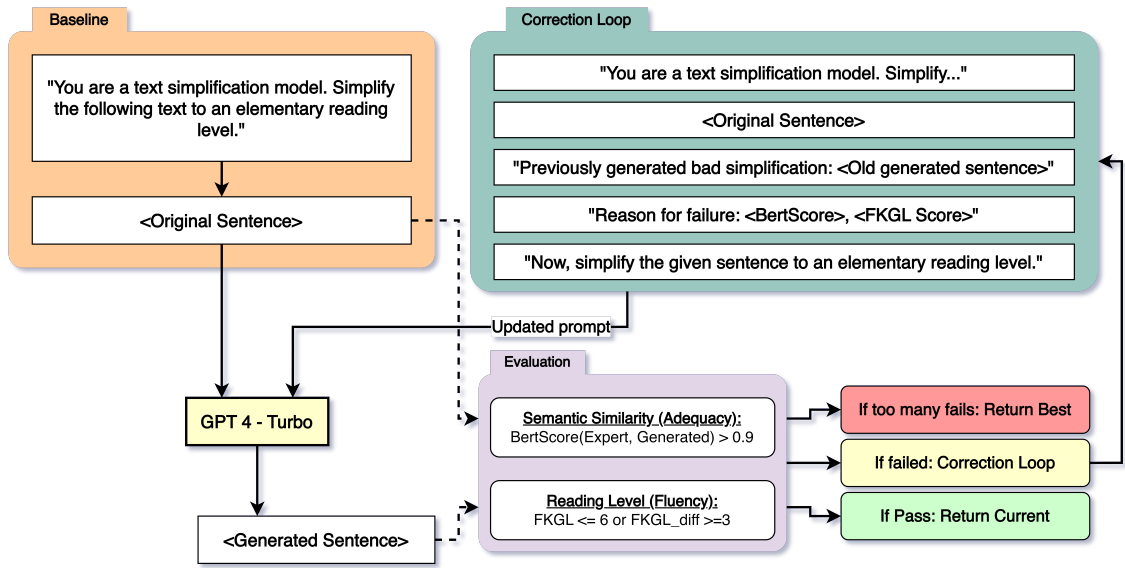}
    \caption{Diagram of Self-correction Prompting}
    \label{sc_fig}
\end{figure}

\textbf{Self-Correction (SC):} Inspired by recent work on self-correction \cite{wu2023self,pan2023automatically}, we created the self-correction approach for  simplification shown in Figure \ref{sc_fig}. The intent was to both exploit the reasoning capabilities of the LLM and incorporate additional external learning signals to iteratively refine the output. Each generated simplification was scored for reading level using FKGL and semantic fidelity using BERTScore. Acceptance criteria include: 1) FKGL score $\le 6$ or FKGL reduced by three grade levels from the original text and 2) BERTScore between the generated and original texts $\ge 0.95$. If the simplification met the acceptance criteria, it was accepted; otherwise, the simplification was fed back into GPT-4 with instructions for improving the simplification. Each sample was processed iteratively at most three times.

\subsubsection{Experimental Paradigm}

We utilized PyTorch \cite{NEURIPS2019_bdbca288} and Hugging Face's Transformers \cite{wolf-etal-2020-transformers}. We used one Llama 2 variant: \textit{Llama-2-7b-chat-hf} \footnote{\url{https://huggingface.co/meta-llama/Llama-2-7b-chat-hf}} and two Llama 3 variants: \textit{Llama-3-8B-Instruct} \footnote{\url{https://huggingface.co/meta-llama/Meta-Llama-3-8B-Instruct}} and \textit{Llama-3-70B-Instruct} \footnote{\url{https://huggingface.co/meta-llama/Meta-Llama-3-70B-Instruct}}. Experimentations were performed using NVIDIA A100 GPUs.

For the \textit{Llama-2-7b-chat-hf} model, all experiments, including zero-shot, SFT, and RL/RLHF, were conducted using a single A100-80GB GPU. For the \textit{Llama-3-8B-Instruct} model, zero-shot and SFT experiments were also feasible using a single A100 GPU; however, RL/RLHF experimentation required two A100-80GB GPUs. For the \textit{Llama-3-70B-Instruct} model, zero-shot inference was conducted using four A100-80GB GPUs without quantization; however, SFT required 4-bit quantization \cite{jin2024comprehensive} with four A100-80GB GPUs. Unfortunately, RL/RLHF experimentation with  \textit{Llama-3-70B-Instruct} was not feasible with our computational resources, even with the 4-bit quantization and utilization of four A100-80GB GPUs.

We also used Weights and Biases \cite{wandb} for tracking training and performance. For SFT, we utilized LORA \cite{hu2021lora} utilities from the Parameter-Efficient Fine-Tuning (PEFT) package \cite{xu2023parameter, peft}, to reduce the time and space complexity of model training. Hyperparameters were selected based on computational constraints and performance optimization. We selected a batch size of 8 to balance memory usage and model updating. The learning rate was set to 2e-4, following preliminary experiments to find a balance between catastrophic forgetting and suboptimal convergence. Additional hyperparameters, such as weight decay (0.001) and maximum gradient norm (0.3), were included to prevent overfitting and maintain the stability of the training process. For RLHF, we specified a range of hyperparameters, including a learning rate of 1.41e-5 and PPO epochs of 2, reflecting a conservative approach to incremental learning. Our choice of batch and mini-batch sizes, set to 4, was influenced by the need to manage the computational load effectively while still capturing the nuances of gradient updates during training. 

\subsection{Evaluation}

We utilized widely accepted evaluation metrics, each addressing specific aspects of text simplification quality, including:

\begin{enumerate}
    \item \textbf{System output Against References and against the Input sentence (SARI)} \cite{xu-etal-2016-optimizing}: SARI evaluates simplification quality by comparing system-generated output with both the original and reference texts. Outputs are assessed based the operations: \textit{add} - words that are in the reference but not in the original text; \textit{keep} - words from the original text are also in the reference, and \textit{delete} - words from the original text that are not in the reference.
   
    \item \textbf{Bilingual Evaluation Understudy (BLEU)} \cite{Papineni02bleu:a, lin-och-2004-orange}: BLEU measures the precision of n-grams in the generated text against a reference by evaluating how many words in the generated text are also present in the reference text. We used 4-gram BLEU.
    \item \textbf{BERTScore} \cite{bert-score}: BERTScore quantifies the semantic similarity between texts using contextual embeddings from the BERT model. It compares sentence-level embeddings of the generated text and reference texts, providing a measure of semantic comparison beyond surface-level text matching.
    \item \textbf{Recall-Oriented Understudy for Gisting Evaluation (ROUGE)} \cite{lin-2004-rouge}: ROUGE focuses on n-gram overlap between the output and reference texts. ROUGE-N measures the overlap of n-grams, and ROUGE-L measures the longest common subsequence. We used ROUGE-L because it is less sensitive to minor differences in wording and is particularly useful for assessing fluency and coherency.
    \item \textbf{Flesch-Kincaid Grade Level (FKGL)} \cite{kincaid1975derivation}: FKGL assesses the reading level of text based on the average number of syllables per word and number of words per sentence. FKGL scores indicate the estimated grade level.
\end{enumerate}

\ifsubfile
\bibliography{mybib}
\fi

\section{Results}

\begin{table}[ht!]
\small
\centering
\begin{tabular}{llcccccc}
\toprule
\multicolumn{8}{c}{\textbf{In-Domain}} \\
\midrule
\textbf{Model Name} & \textbf{Variant} & \multicolumn{1}{l}{\textbf{\begin{tabular}[c]{@{}l@{}}
Exp.\\ No.\end{tabular}}} & \multicolumn{1}{l}{\textbf{\begin{tabular}[c]{@{}l@{}}
SARI\\ ($\uparrow$)\end{tabular}}} & \multicolumn{1}{l}{\textbf{\begin{tabular}[c]{@{}l@{}}
BLEU\\ ($\uparrow$)\end{tabular}}} & \multicolumn{1}{l}{\textbf{\begin{tabular}[c]{@{}l@{}}
BERT\\ Score ($\uparrow$)\end{tabular}}} & \multicolumn{1}{l}{\textbf{\begin{tabular}[c]{@{}l@{}}
ROUGE\\ ($\uparrow$)\end{tabular}}} & \multicolumn{1}{l}{\textbf{\begin{tabular}[c]{@{}l@{}}
FKGL\\ ($\downarrow$)\end{tabular}}} \\
\midrule
\multirow{2}{*}{\begin{tabular}[c]{@{}l@{}}Llama-2-7b-\\ chat-hf\end{tabular}} & Zero-shot & 01 & 28.43 & 0.15 & 0.90 & 0.48 & 9.08 \\   
 & SFT & 02 & 47.92 & 0.47 & 0.95 & 0.70 & 7.08 \\ 
\midrule
\multirow{2}{*}{\begin{tabular}[c]{@{}l@{}}Llama-3-\\8B-Instruct\end{tabular}} & Zero-shot & 03 & 25.15 & 0.11 & 0.90 & 0.46 & 8.96 \\ 
 & SFT & 04 & 52.04 & 0.42 & 0.93 & 0.61 & 6.30 \\ 
\midrule
\multirow{2}{*}{\begin{tabular}[c]{@{}l@{}}Llama-3-\\70B-Instruct\end{tabular}} & Zero-shot & 05 & 24.49 & 0.02 & 0.84 & 0.12 & 8.59 \\ 
 & SFT & 06 & 46.73 & 0.39 & 0.93 & 0.56 & 5.16 \\ 
\midrule
\multirow{4}{*}{\begin{tabular}[c]{@{}l@{}}Llama-2-7b-\\ chat-hf\end{tabular}} & RL ($R_{FKGL+Rel}$) & 07 & 29.99 & 0.29 & 0.91 & 0.54 & 8.29 \\ 
 & RLHF ($R_{OvS}$) & 08 & 31.62 & 0.30 & 0.91 & 0.56 & 8.40 \\ 
 & RLHF ($R_{FKGL+OvS}$) & 09 & 29.03 & 0.28 & 0.91 & 0.54 & 8.38 \\ 
 & SFT+RLHF ($R_{FKGL+OvS}$) & 10 & 60.39 & 0.70 & 0.97 & 0.89 & 8.24 \\ 
\midrule
\multirow{4}{*}{\begin{tabular}[c]{@{}l@{}}Llama-3-\\8B-Instruct\end{tabular}} & RL ($R_{FKGL+Rel}$) & 11 & 20.44 & 0.11 & 0.88 & 0.32 & 7.38 \\ 
 & RLHF ($R_{OvS}$) & 12 & 42.45 & 0.37 & 0.91 & 0.51 & 8.02 \\ 
 & RLHF ($R_{FKGL+OvS}$) & 13 & 55.01 & 0.59 & 0.93 & 0.80 & 8.24 \\ 
 & SFT+RLHF ($R_{FKGL+OvS}$) & 14 & 50.98 & 0.53 & 0.91 & 0.73 & 8.11 \\ 
\midrule
\multirow{5}{*}{GPT-4 Turbo} & Zero-shot & 15 & 22.03 & 0.14 & 0.92 & 0.43 & 5.71 \\
 & In-context & 16 & 22.04 & 0.14 & 0.92 & 0.42 & 5.68 \\
 & FKGL-Enhanced & 17 & 23.78 & 0.24 & 0.94 & 0.53 & 5.37 \\
 & CoT & 18 & 20.97 & 0.18 & 0.93 & 0.46 & 4.71 \\
 & Self-Correction & 19 & 22.86 & 0.21 & 0.93 & 0.51 & 4.84 \\
\bottomrule
\toprule
\multicolumn{8}{c}{\textbf{Domain Adaptation}} \\
\midrule
\textbf{Model Name} & \textbf{Variant} & \multicolumn{1}{l}{\textbf{\begin{tabular}[c]{@{}l@{}}
Exp.\\ No.\end{tabular}}} & \multicolumn{1}{l}{\textbf{\begin{tabular}[c]{@{}l@{}}
SARI\\ ($\uparrow$)\end{tabular}}} & \multicolumn{1}{l}{\textbf{\begin{tabular}[c]{@{}l@{}}
BLEU\\ ($\uparrow$)\end{tabular}}} & \multicolumn{1}{l}{\textbf{\begin{tabular}[c]{@{}l@{}}
BERT\\ Score ($\uparrow$)\end{tabular}}} & \multicolumn{1}{l}{\textbf{\begin{tabular}[c]{@{}l@{}}
ROUGE\\ ($\uparrow$)\end{tabular}}} & \multicolumn{1}{l}{\textbf{\begin{tabular}[c]{@{}l@{}}
FKGL\\ ($\downarrow$)\end{tabular}}} \\
\midrule
\multirow{3}{*}{\begin{tabular}[c]{@{}l@{}}Llama-2-7b-\\ chat-hf\end{tabular}} & none & 20 & 35.15 & 0.52 & 0.94 & 0.64 & 8.49 \\ 
 & $R_{FKGL}+R_{Rel}$ & 21 & 45.71 & 0.63 & 0.96 & 0.78 & 7.50 \\ 
 & $R_{FKGL}+R_{OvS}$ & 22 & 74.41 & 0.73 & 0.97 & 0.89 & 7.74 \\ 
\midrule
\multirow{3}{*}{\begin{tabular}[c]{@{}l@{}}Llama-3-\\8B-Instruct\end{tabular}} & none & 23 & 38.73 & 0.43 & 0.93 & 0.60 & 6.36 \\ 
 & $R_{FKGL}+R_{Rel}$ & 24 & 49.97 & 0.45 & 0.92 & 0.59 & 7.27 \\ 
 & $R_{FKGL}+R_{OvS}$ & 25 & 51.06 & 0.57 & 0.91 & 0.72 & 8.39 \\ 
\bottomrule
\end{tabular}
\caption{In-domain and domain adaptation performance of Prompt-based, SFT, and RLHF experimentation on SimpleDC dataset. Source FKGL: 8.11, Target FKGL: 7.65. Higher scores indicate better performance for SARI, BLEU, BERTScore, and ROUGE ($\uparrow$), whereas lower scores are preferred for FKGL ($\downarrow$).}
\label{text_simplification}
\end{table}

\subsection{Text Simplification}
Table \ref{text_simplification} presents the performance on the withheld SimpleDC test set, evaluating: \textit{Llama-2-7b-chat-hf}, \textit{Llama-3-8B-Instruct}, \textit{Llama-3-70B-Instruct}, and GPT-4 Turbo. In the zero-shot setting \textit{(Exp. 01, 03, 05, and 15)}, \textit{Llama-2-7b-chat-hf} achieved the highest SARI score, although GPT-4 achieved the lowest FKGL. SFT with the Llama models \textit{(Exp. 02, 04, and 06)} markedly improved performance across SARI, BLEU, BERTScore, and ROUGE metrics, while reducing the FKGL by approximately two grade levels. These results underscore the role of SFT in understanding and learning the SimpleDC annotation schema and the annotators' stylistic nuances. Notably, both \textit{Llama-2-7b-chat-hf} and \textit{Llama-3-8B-Instruct} showed substantial performance improvements with SFT, with 47.92 and 52.04 SARI and 7.08 and 6.32 FKGL, respectively, indicating effective task training. \textit{Llama-3-70B-Instruct} also demonstrated noteworthy improvements with SFT \textit{(Exp. 06)}, achieving 46.73 SARI and the lowest reading level at 5.16 FKGL.

The success of the RL/RLHF varies by Llama variant and reward strategy. In RL/RLHF-only experiments, \textit{Llama-3-8B-Instruct} \textit{(Exp. 11-13)} consistently outperformed \textit{Llama-2-7b-chat-hf} \textit{(Exp. 07-09)} across multiple metrics, achieving higher SARI and lower FKGL. For\textit{Llama-3-8B-Instruct}, the proposed RLHF reward function, $R_{FKGL+OvS}$, markedly increased the SARI score from 20.44 \textit{(Exp. 11)} in the RL scenario to 55.01 \textit{(Exp. 13)}. These findings emphasize the importance of integrating human feedback from the target task/data set through RLHF, as compared to the data set agnostic feedback of RL alone.

When combining SFT with RLHF for \textit{Llama-2-7b-chat-hf} \textit{(Exp. 10)}, the model achieved the highest performance, at 60.39 SARI, 0.70 BLEU, and 0.97 BERTScore. This configuration also yielded a relatively low reading level at 8.24 FKGL, demonstrating that SFT and RLHF can be complementary in enhancing text simplification capabilities. However, for \textit{Llama-3-8B-Instruct}, combining SFT with RLHF \textit{(Exp. 14)} also showed no discernable improvement over RLHF alone \textit{(Exp. 13)}. In this configuration, the model achieved a SARI of 50.98, BLEU of 0.53, BERTScore of 0.91, and FKGL of 8.11.

For in-context learning, GPT-4 Turbo \textit{(Exp. 15-19)} achieved lower scores across all adequacy metrics (SARI, BLEU, BERTScore, and ROUGE) compared to the Llama models \textit{(Exp. 01-14)} but achieved the lowest FKGL. Lower performance does not necessarily indicate lower quality simplifications generated by GPT-4. Based on our qualitative review, GPT-4 edited the original text more extensively, resulting in a larger divergence between the original and simplified text. Additionally, the voice or writing style of GPT-4 appears to differ from that of our annotators, contributing to the lower performance for adequacy metrics. The FKGL-Enhanced prompting strategy \textit{(Exp. 17)} achieved the best performance across adequacy metrics, and the CoT prompting strategy \textit{(Exp. 18)} achieved the lowest FKGL. The Self-Correction approach \textit{(Exp. 19)} achieved a balance between adequacy and reading level at 22.86 SARI and 4.48 FKGL.

\subsection{Domain Adaptation}

We used the Med-EASi corpus to explore the role of RL/RLHF in domain adaptation by first conducting out-of-domain SFT with Med-EASi, followed by in-domain RL/RLHF training using unlabeled digestive cancer text. Table \ref{text_simplification} \textit{(Exp. 20-25)} presents the domain adaptation results. This experimentation evaluated the feasibility of leveraging out-of-domain corpora to enhance in-domain performance. In this domain adaptation setting, the performance gains from RL/RLHF vary by Llama variant and specific reward function. For \textit{Llama-3-8B-Instruct}, continued in-domain RLHF using the $R_{FKGL+OvS}$ reward markedly improved performance over out-of-domain SFT on Med-EASi alone, increasing SARI from 38.73 to 51.06, BLEU from 0.43 to 0.57, and ROUGE from 0.60 to 0.72 \textit{(Exp. 23 vs. Exp. 25)}. However, \textit{Llama-3-8B-Instruct} with RLHF alone and no prior SFT on Med-EASI \textit{(Exp. 13)} achieved better results overall, indicating the combined out-of-domain SFT and in-domain RLHF training were not complementary.

For \textit{Llama-2-7b-chat-hf}, RL/RLHF alone \textit{(Exp. 07-09)} was not as effective; however, the best performance across all models was achieved by combing out-of-domain SFT on Med-EASi with in-domain RLHF with the $R_{FKGL+OvS}$ reward. For this configuration, continued training through in-domain RLHF increased SARI from 35.15  to 74.41, BLEU from 0.52 to 0.73, and ROUGE from 0.64 to 0.89 and decreased FKGL from 8.49 to 7.74 \textit{(Exp. 20 vs Exp. 22)}. The $R_{FKGL+Rel}$ reward \textit{(Exp. 21)} also led to improvements over out-of-domain SFT alone but was less effective compared to the $R_{FKGL+OvS}$ reward. The domain adaptation results highlight that the top-performing domain-adapted LLM \textit{(Exp. 22)} achieves comparable performance to the best in-domain models \textit{(Exp. 10, 13, 04 and 14)}. These findings affirm the potential of combining SFT with RL/RLHF to enhance performance while minimizing annotation data requirements and associated costs.

\subsection{Human Evaluation}

\begin{table}[ht!]
\small
\centering
\begin{tabular}{l|cccc|cccc}
\toprule
\multicolumn{9}{c}{\textbf{Preservation of Meaning}} \\
\midrule
\textbf{} & \multicolumn{4}{c}{\textbf{Human vs Llama 2}} & \multicolumn{4}{|c}{\textbf{Human vs GPT-4}} \\
\midrule
Source & Human & AI & Both & None & Human & AI & Both & None \\
\midrule
CDC & 0 & 4.35 & 95.65 & 0     & 62.32 & 10.14 & 21.74 & 5.80 \\
NCI & 8.89 & 12.22 & 78.89 & 0 & 56.67 & 15.56 & 27.78 & 0    \\
ACS & 11.11 & 7.19 & 81.70 & 0 & 50.98 & 6.54 & 41.18 & 1.31  \\
\midrule
Overall & 8.01 & 8.01 & 83.97 & 0 & 55.13 & 9.94 & 33.02 & 1.93  \\
\bottomrule
\toprule
\multicolumn{9}{c}{\textbf{Understandability}} \\
\midrule
\textbf{} & \multicolumn{4}{c}{\textbf{Human vs Llama 2}} & \multicolumn{4}{|c}{\textbf{Human vs GPT-4}} \\
\midrule
Source & Human & AI & Both & None & Human & AI & Both & None \\
\midrule
CDC & 18.84 & 1.45 & 75.36 & 4.35  & 43.48 & 42.03 & 8.70 & 5.80   \\
NCI & 30 & 17.78 & 46.67 & 5.56    & 35.56 & 48.89 & 14.44 & 1.11  \\
ACS & 19.61 & 10.46 & 65.36 & 4.58 & 35.95 & 56.86 & 5.23 & 1.96   \\
\midrule
Overall & 22.4 & 10.6 & 62.2 & 4.8 & 37.5 & 51.3 & 8.7 & 2.6  \\
\bottomrule
\end{tabular}
\caption{Human evaluation of Human-generated simplification and AI-generated simplification}
\label{human-vs-ai}
\end{table}

The above assessment focuses on adequacy and fluency, where adequacy refers to the preservation of the original meaning, and fluency assesses the readability of the simplification. Automatic adequacy and fluency metrics provide useful information; however, these metrics do not holistically assess the quality of simplifications \cite{kauchak2016moving}. To provide a comprehensive evaluation, we manually evaluated human- and AI-generated simplifications. Manual evaluations were performed by three nurse practitioners with extensive knowledge of patient care and education, who were not involved in the SimpleDC annotation. The nurse practitioners reviewed a subset of the test samples in an \textit{A-B} comparison, to evaluate the preservation of medical meaning and understandability for a broad audience. In this A-B comparison, the reviewers could indicate their preference for \textit{A}, \textit{B}, \textit{Both}, or \textit{None}. The reviewers were blinded to which samples, A or B, were human- or AI-generated, and the order (A vs. B assignments) was shuffled. The evaluation included: 1) human vs. the best performing Llama (\textit{Exp. 22}) and 2) human vs. the best performing GPT-4 (\textit{Exp. 19}). The manual evaluation comprised a total of 104 sentence pairs, including 23 from CDC, 30 from NCI, and 51 from ACS.

Table \ref{human-vs-ai} presents the results of the human evaluation, which demonstrate human and Llama simplifications consistently preserve the meaning of the original text and use understandable language. This further validates the results in Table \ref{text_simplification}, specifically that the trained Llama model successfully learns the simplification task and emulates human simplification. The reviewers preferred human simplifications over GPT-4 simplifications to preserve meaning. The reviewers preferred the readability of GPT-4 over human simplification, suggesting a trade-off between fluency and adherence to the original content. This nuanced evaluation highlights the complexity of balancing adequacy and fluency in text simplification tasks, underscoring the need for a multifaceted approach in assessing model performance.

\subsection{Error Analysis}

\begin{table}[ht!]
\small
\centering
\begin{tabular}{l|ccc|ccc|ccc}
\toprule
 & \multicolumn{3}{c}{\textbf{Unchanged from   original}} & \multicolumn{3}{|c}{\textbf{Insertions in new}} & \multicolumn{3}{|c}{\textbf{Deletions from   original}} \\
 \midrule
\multicolumn{1}{c}{Source} & \multicolumn{1}{|c}{Human} & \multicolumn{1}{c}{Llama 2} & \multicolumn{1}{c}{GPT-4} & \multicolumn{1}{|c}{Human} & \multicolumn{1}{c}{Llama 2} & \multicolumn{1}{c}{GPT-4} & \multicolumn{1}{|c}{Human} & \multicolumn{1}{c}{Llama 2} & \multicolumn{1}{c}{GPT-4} \\
\midrule
CDC & 92.1\% & 98.2\% & 53.9\% &  9.3\% &  1.3\% & 57.3\% &  7.9\% &  1.8\% & 46.1\% \\
NCI & 72.8\% & 81.2\% & 41.5\% & 24.7\% & 13.2\% & 65.5\% & 27.2\% & 18.8\% & 58.5\% \\
ACS & 96.4\% & 92.9\% & 51.9\% &  2.9\% &  3.6\% & 60.8\% &  3.6\% &  7.1\% & 48.1\% \\
\midrule
Overall & 88.6\% & 90.7\% & 49.3\% & 10.6\% & 5.9\% & 61.4\% & 11.4\% & 9.3\% & 50.7\% \\
\bottomrule
\end{tabular}
\caption{Error Analysis of the best Llama 2 model (Exp 22) and the best GPT-4 model (Exp 19)}
\label{error-analysis}
\end{table}

Despite GPT-4's advanced language capabilities and significantly larger model size, the trained Llama models outperformed GPT-4 in both automated metrics and manual evaluations. This unexpected result can be attributed to differences in task interpretation. The simplification task was designed to improve the accessibility of the text for patients without altering its meaning. To better understand the editing performed by human annotators and LLMs to create the simplified versions, we calculated the proportion of tokens that are unchanged (not edited), inserted, and deleted from the original text. Table \ref{error-analysis} presents a summary of the unchanged, inserted, and deleted words in the SimpleDC test set, normalized by the number of tokens in the original text (comparing with the Llama 2 model outputs from \textit{Exp. 22}). In creating SimpleDC, the annotators edited the NCI content more heavily than the ACS and CDC content because the NCI content started at a higher reading grade level (median 7.6 for ACS, 7.0 for CDC, and 8.3 for NCI). The Llama 2 model performed very light editing, deleting 9.3\% of existing tokens and inserting 5.9\% new tokens. In contrast, GPT-4 made many more edits, deleting 50.7\% of existing tokens and inserting 61.4\% new tokens. This discrepancy highlights the importance of task alignment between human expectations and AI model interpretations.

The GPT-4 performance was negatively impacted in the automatic and manual evaluations by over-generation. This tendency to add new text contrasts sharply with the more conservative approach of human annotators and the Llama models, which introduced new content at rates of approximately 10.6\% and 5.9\% respectively, with average word counts of 14.24 and 14.05. Although these insertions can enhance understandability by adding explanatory content or simplifying complex ideas, they can also stray from the original message. For example, the GPT model simplified the sentence ``The main functions of the liver include the following:" into a 30+ word sentence (``The liver has some big jobs ... help break down fat in food."). While this type of generation may be helpful for the audience, it does not align with the objective of the task.

The analysis underscores the differing abilities of the models to adapt to the inherent reading complexity of the source materials, notably between human annotators and trained Llama 2 and GPT-4 models. Specifically, both human annotators and the trained Llama 2 model demonstrated a nuanced ability to modulate their simplification efforts in response to the reading level complexity of the sources (ACS, NCI, and CDC). This adaptability is indicative of a strategic approach to text simplification, where the extent of modifications is calibrated based on the initial complexity of the source material. Conversely, the GPT-4 strategy appeared relatively uniform across sources, indicating a lack of sensitivity to the nuanced demands of the simplification task. Such uniformity may imply a limitation in prompt-based approaches to simplification.

\ifsubfile
\bibliography{mybib}
\fi

\section{Discussion}

Our research explores the effectiveness of LLMs for simplifying educational health information, contrasting the performance of SFT, RL, RLHF, and in-context learning approaches. We explore the role of model training for specific tasks, revealing how GPT models, despite their general prowess, require targeted adjustments to excel in specialized contexts. We introduce a novel reward model designed to optimize simplification by focusing on readability, relevance, and fidelity to the original text's intent. This new model incorporates an automatic reading-level metric (FKGL) and the innovative Original vs. Simplified (OvS) reward informed by human feedback. Our findings demonstrate how RLHF can improve the quality of text simplifications, making complex medical information more accessible and understandable to patients. The integration of human feedback into the reinforcement learning process, through the OvS reward, marks a significant advancement in aligning AI-generated content with human expectations. Our study advances AI-driven text simplification for healthcare communication and demonstrates the potential for task-specific LLMs to bridge the gap between medical expertise and patient understanding.

Our results indicate that foundational LLMs, with their unique architectures and training protocols, respond differently to continued training. Specifically, Llama-2 and Llama-3 exhibited varied responses to SFT and RLHF. For Llama-2, SFT improved performance across all metrics compared to zero-shot; however, RLHF did not significantly alter performance relative to zero-shot. Combining SFT and RLHF for Llama-2 resulted in improved performance relative to either method used individually in both in-domain and domain adaptation experiments. In contrast, for Llama-3, both SFT and RLHF independently improved performance, with in-domain RLHF training yielding the best results. However, combining SFT and RLHF for Llama-3 did not enhance performance beyond that of RLHF alone. The architectural and training specifics for Llama-3 are not yet available, making it difficult to fully understand the differences between Llama-2 and Llama-3 and the reasons for these performance discrepancies.

The ability to accurately simplify complex medical texts without losing essential information is crucial for improving patient understanding and engagement. Our research illustrates the potential of using AI to assist in this process but also highlights the challenges in ensuring the simplifications are appropriate and accessible to all patients. This emphasizes the need for a collaborative approach that combines technical innovation with input from healthcare professionals to ensure the simplifications meet the diverse needs of the patient population. By leveraging AI's capabilities alongside expert human feedback, we can better tailor health information to enhance patient comprehension and empowerment.

\ifsubfile
\bibliography{mybib}
\fi

\section{Conclusions}

In this paper, we present SimpleDC, a carefully curated human-annotated text simplification corpus for digestive cancer education. The rich annotations of the dataset have been instrumental in training effective simplification models. We explore a range of learning approaches, including SFT, RL, RLHF, and prompt-based approaches. We introduced a new RLHF reward function, $R_{FKGL+OvS}$, which outperforms an existing RL reward function for text simplification. In terms of model performance, we observed that Llama models with task-specific training outperformed GPT-4. This highlights the impact of high-quality task-specific annotated data in improving model performance and simplification quality, including balancing comprehensibility and medical accuracy. Our results also demonstrate that RL/RLHF can be used to incorporate unlabeled text into training to improve simplification performance: 1) as a standalone training step, 2) in conjunction with SFT, and 3) as a domain adaptation technique. Although this represents a significant advancement, our findings also underscore the vast potential for future research in this area. Our work provides a robust foundation for further explorations, aiming to refine and expand the capabilities of NLP models to simplify complex medical information for enhanced patient understanding. Future research could explore integrating multimodal data sources, such as medical images or patient educational videos, alongside textual information to further enhance the simplification models' ability to provide comprehensive and accessible health education. Additionally, investigating the incorporation of feedback mechanisms from actual patients and healthcare professionals into the model training process could significantly improve the relevance and usability of simplified medical content.

\ifsubfile
\bibliography{mybib}
\fi

\section{Data Availability and Legal Considerations}

We provide the new SimpleDC corpus to the research community to advance health text simplification.\footnote{\href{https://github.com/mushfiqur11/simpledc-dataset.git}{https://github.com/mushfiqur11/simpledc-dataset.git}} In developing SimpleDC, we utilized publicly available text data from the NCI, ACS, and CDC websites with the goal of advancing text simplification and public health communication. Our use of this data is in alignment with fair use principles, as the data was employed exclusively for academic and research purposes, without any commercial intent. We minimized data usage, extracting only the portions necessary to support our work on text simplification and public health communication. Furthermore, our use of the data does not adversely affect the market value or potential of the original content. The text data from the NCI, ACS, and CDC was specifically used to improve NLP methods that enhance the accessibility of health information, with all sources carefully attributed. SimpleDC is made publicly available, with detailed source attribution, ensuring full transparency and traceability.


\section*{Declaration of Generative AI and AI-assisted Technology in the Scientific Writing Process}
During the preparation of this work, the author(s) used ChatGPT and Writefull to solicit editorial feedback regarding writing clarity and proofreading. All the scientific content and data interpretation remained solely the authors' contributions. After using these tools/services, the authors reviewed and edited the content as needed and take full responsibility for the content of the publication.

\bibliography{mybib,Kai_LexicalSimp}

\appendix
\section{Case Study of Simplification Outputs}

\begin{table}[ht!]
\small
\centering
\begin{tabular}{llp{9cm}}
\toprule
\textbf{Source/Model} & \multicolumn{1}{l}{\textbf{\begin{tabular}[c]{@{}l@{}}Exp.\\ No.\end{tabular}}} & \multicolumn{1}{l}{\textbf{\begin{tabular}[c]{@{}l@{}}Simplification\\Output\end{tabular}}} \\
\midrule
Original & - & As food is digested, it passes from the stomach to the small intestine. \\
Target Simplification & - & As food is broken down in your body, it passes from the stomach to the section of the gut called the small intestine. \\
\midrule
\multirow{2}{*}{\begin{tabular}[c]{@{}l@{}}Llama-2-7b with SFT(SimpleDC) +\\RLHF (SimpleDC)\end{tabular}} & 10 & Food is digested as it passes from the stomach to the small intestine. \\
\multirow{2}{*}{\begin{tabular}[c]{@{}l@{}}Llama-2-7b with SFT(MedEASi) +\\RLHF (SimpleDC)\end{tabular}} & 22 & At this point, food passes from the stomach to the colon. \\ \\
\midrule
\multirow{2}{*}{\begin{tabular}[c]{@{}l@{}}Llama-3-8b with SFT(SimpleDC) +\\RLHF (SimpleDC)\end{tabular}} & 14 & Food is digested and it passes from the stomach to the small intestine. \\
\multirow{2}{*}{\begin{tabular}[c]{@{}l@{}}Llama-3-8b with SFT(MedEASi) +\\RLHF (SimpleDC)\end{tabular}} & 25 & As food is digested, it goes from the stomach to the small intestine. \\
\midrule
Llama-3-70b with SFT(SimpleDC) & 06 & When you eat food, it goes from the stomach to the small intestine. This is called the small bowel. It is like a long, thin, winding road. \\
\midrule
GPT-4 with Self-correction & 19 & When we eat, the food goes from our stomach to our small intestine. \\
\bottomrule

\midrule
Original & - & Basal cell carcinomas are a type of skin cancer that can develop in the perianal skin. \\
Target Simplification & - & Basal cell cancer is a kind of skin cancer that can sometimes start in the perianal skin. \\
\midrule
\multirow{2}{*}{\begin{tabular}[c]{@{}l@{}}Llama-2-7b with SFT(SimpleDC) +\\RLHF (SimpleDC)\end{tabular}} & 10 & Basal cell carcinomas are a type of skin cancer that can develop in the perianal skin.  \\
\multirow{2}{*}{\begin{tabular}[c]{@{}l@{}}Llama-2-7b with SFT(MedEASi) +\\RLHF (SimpleDC)\end{tabular}} & 22 & Basal cell cancer is a type of skin cancer that can develop in the perianal skin.\\
\midrule
\multirow{2}{*}{\begin{tabular}[c]{@{}l@{}}Llama-3-8b with SFT(SimpleDC) +\\RLHF (SimpleDC)\end{tabular}} & 14 & Basal cell cancers are a type of skin cancer that can develop in the perianal skin.  \\
\multirow{2}{*}{\begin{tabular}[c]{@{}l@{}}Llama-3-8b with SFT(MedEASi) +\\RLHF (SimpleDC)\end{tabular}} & 25 & Basal cell carcinomas are a type of skin cancer that can develop in the perianal skin. \\
\midrule
Llama-3-70b with SFT(SimpleDC) & 06 & Basal cell carcinomas are a kind of skin cancer that can grow in the perianal skin. This is called anal melanosis. \\
\midrule
GPT-4 with Self-correction & 19 & Basal cell carcinomas are a kind of skin cancer. They can grow on the skin around the bottom. \\
\bottomrule
\end{tabular}
\caption{Simplification outputs of the same sentence by different models.}
\label{tab:case_study}
\end{table}

Table \ref{tab:case_study} presents text simplification examples from Llama-2 and Llama-3 models using various fine-tuning and reinforcement learning approaches. Both Llama-2-7b and Llama-3-8b models generally produce concise and accurate simplifications. However, the Llama-3-70b model, while generating seemingly more readable outputs, tends to add additional information not present in the original text. These results underscore the importance of model choice and configuration in text simplification, highlighting the need for precise model selection and fine-tuning in domain-specific applications.



\end{document}